\documentclass[runningheads]{llncs}

\usepackage{eccv}

\usepackage{eccvabbrv}

\usepackage{graphicx}
\usepackage{booktabs}

\usepackage[accsupp]{axessibility}  

\usepackage[pagebackref,breaklinks,colorlinks,citecolor=eccvblue]{hyperref}

\usepackage{orcidlink}

\usepackage{algorithm}
\usepackage{algpseudocode}
\usepackage{threeparttable}
\usepackage[T1]{fontenc}
\usepackage{color}
\usepackage{xcolor}
\usepackage{multirow}
\usepackage{overpic}
\usepackage{wrapfig}
\usepackage{bbm}
\usepackage{marvosym}

\usepackage{enumitem}
\usepackage{listings}

\usepackage{tcolorbox}
\tcbuselibrary{listings, breakable, skins}

\newtcblisting{promptbox}[1]{
    enhanced,                  
    breakable,                 
    colback=gray!5!white,      
    colframe=gray!60!black,    
    coltitle=black,            
    colbacktitle=gray!15!white,
    fonttitle=\bfseries\small, 
    title=#1,                  
    boxrule=0.5pt,             
    arc=3pt,                   
    left=4pt, right=4pt,       
    top=4pt, bottom=4pt,       
    listing only,              
    listing options={
        basicstyle=\small\ttfamily, 
        breaklines=true,            
        breakatwhitespace=true,     
        columns=fullflexible,       
        keepspaces=true,
    }
}

\begin{document}

\title{Bridging VideoQA and Video-Guided Agentic Tasks via Generalized Keyframe Extraction}

\titlerunning{VG-GUI-Bench and TASKER}

\author{Sunqi Fan\orcidlink{0009-0003-0777-9408} \and
Qingle Liu\orcidlink{0009-0003-6764-3094} \and
Runqi Yin\orcidlink{0009-0000-5981-9801}  \and
Meng-Hao Guo\orcidlink{0000-0002-4128-4594} \and 
Shuojin Yang\textsuperscript{\Letter}\orcidlink{0009-0009-5345-7972} 
}


\authorrunning{S. Fan et al.}

\institute{Tsinghua University, Beijing, China\\
\email{\{fansq20, lql24, yrq24\}@mails.tsinghua.edu.cn, \\\{gmh, yangshuojin\}@tsinghua.edu.cn}\\
Project Page: \url{https://vg-gui-tasker.github.io/}
}


\maketitle

{\renewcommand{\thefootnote}{}\footnotetext{\Letter~Corresponding author.}}

\begin{abstract}

Video understanding is a fundamental capability for multimodal intelligence, and recent Multimodal Large Language Models (MLLMs) have achieved remarkable performance on Video Question Answering (VideoQA) benchmarks.
However, existing benchmarks primarily evaluate whether models can perceive shallow visual cues, while rarely examining whether MLLMs can learn deeper knowledge or procedural skills from video tutorials and generalize them to downstream long-horizon agentic tasks.
To address this gap, we introduce \textbf{VG-GUI-Bench} (Video-Guided GUI Benchmark), a new benchmark designed to evaluate whether MLLM-based GUI agents can follow video tutorials to complete corresponding GUI interactive tasks. 
Furthermore, we observe that the performance of models on both VideoQA and video-guided agentic tasks critically depends on effective keyframe extraction. 
Based on this observation, we propose \textbf{TASKER} (Task-driven And Scene-aware Keyframe searchER), a keyframe extraction algorithm that jointly considers task relevance and scene dynamics to identify informative frames. 
Experimental results demonstrate that TASKER achieves significant performance improvements on both VideoQA and video-guided agentic task benchmarks, outperforming the best baseline by 2.0\% on the EgoSchema fullset and 1.8\% on the NExT-QA dataset, respectively. These results further highlight the potential of generalized keyframe extraction methods for video understanding tasks.
Our code and data are available at \url{https://github.com/VG-GUI-TASKER/VG-GUI-TASKER}.

\end{abstract} 
\section{Introduction}
\label{sec:intro}

Multimodal large language models have recently substantially advanced video understanding, achieving strong results on popular VideoQA benchmarks~\cite{fu2024videommefirstevercomprehensiveevaluation, wu2024longvideobenchbenchmarklongcontextinterleaved}. Despite this progress, the current evaluation paradigm remains largely centered on shallow perceptual cues, such as recognizing objects, attributes, and short-term actions. As a consequence, existing benchmarks provide limited insight into a more fundamental question: can MLLMs learn deeper knowledge or procedural skills from videos and generalize them to solve new tasks that require long-horizon agentic capabilities? This limitation becomes particularly evident in real-world learning scenarios. Video tutorials are widely used to teach complex procedures, ranging from software operations to device configuration and daily-life skills. Solving such tasks requires more than recognizing visual events; it involves understanding step-by-step procedures, abstracting key operations, and transferring them to new environments. In other words, models must be able to extract actionable knowledge from videos and apply it to downstream tasks, which is a capability that can be viewed as a form of Video In-Context Learning.

\begin{figure*}[t]
  \centering
  \includegraphics[width=\linewidth]{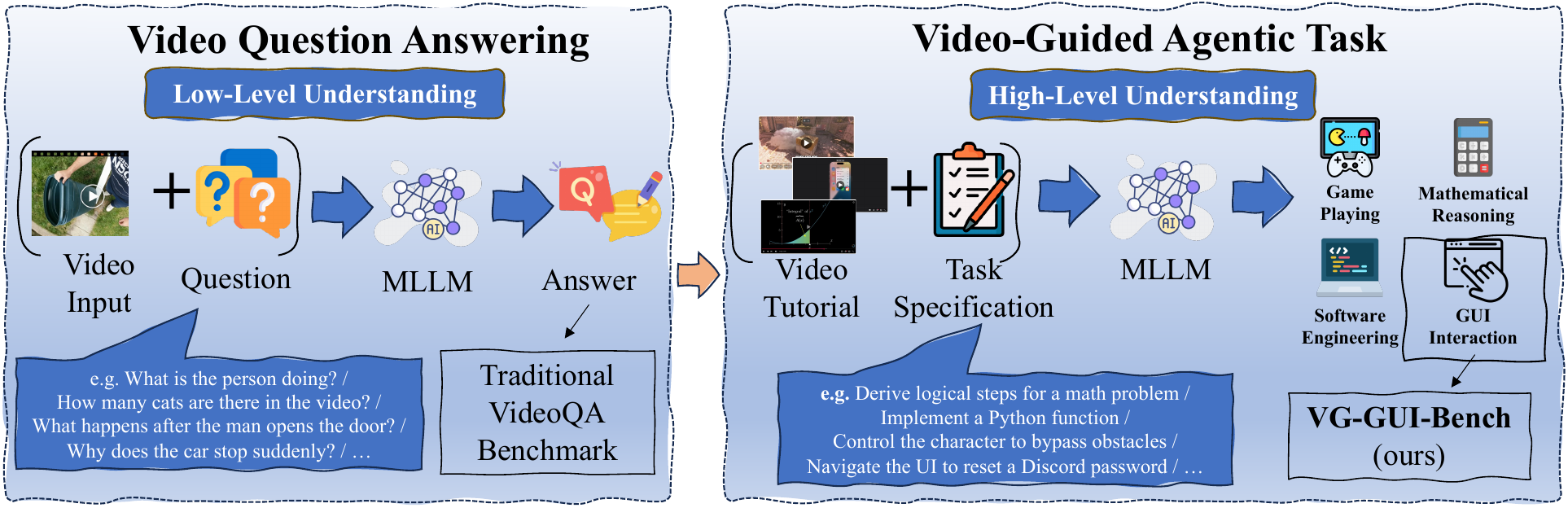}
  \caption{Demonstration of the 2 progressive levels. This work aims to advance video understanding from the VideoQA paradigm (low-level understanding) toward the Video-Guided Agentic Task paradigm (high-level understanding).}
  \label{fig:level}
\end{figure*}

To better understand this capability gap, we characterize video understanding along two progressive levels, forming a natural hierarchy from perception to action, as in Figure \ref{fig:level}:

\begin{itemize}
\item \textbf{Low-Level: VideoQA.}
The foundational task of video understanding, where models must identify temporally relevant moments, comprehend visual evidence, and perform question-conditioned reasoning to extract factual information from videos.

\item \textbf{High-Level: Video-Guided Agentic Tasks.}  
A more advanced setting where models are expected to learn procedural knowledge from video demonstrations and transfer it to long-horizon decision making and execution. For example, a model may watch a tutorial on \emph{“how to change a Discord account password”} and translate the demonstrated procedure into step-by-step actions within a structured Graphical User Interface (GUI) environment.

\end{itemize}


To further evaluate MLLMs' high-level video understanding ability, we introduce \textsc{VG-GUI-Bench} (Video-Guided GUI Benchmark), a benchmark designed to evaluate whether MLLM-based GUI agents can follow video tutorials to complete semantically related tasks. In \textsc{VG-GUI-Bench}, each tutorial video is paired with a corresponding GUI task that requires similar procedural knowledge. This pairing enables systematic evaluation of whether models can extract procedural steps from videos and generalize them to new interactive tasks, providing a practical testbed for studying video in-context learning.

Beyond benchmark design, our study reveals a shared bottleneck across both VideoQA and video-guided agentic tasks: model performance critically depends on how the model identifies and attends to task-relevant temporal content within videos. Long videos often contain redundant or irrelevant segments, while key procedural evidence may appear only briefly. As a result, naive frame sampling strategies either miss critical moments or introduce excessive redundancy, leading to degraded reasoning and inefficient computation. For example, on the NExT-QA benchmark, using GPT-4o with a frame selection strategy improves accuracy by approximately 15\% compared to GPT-4o with uniform sampling, with the same number of frames.


To tackle this challenge, we propose \textsc{TASKER} (Task-driven And Scene-aware Keyframe searchER), a generalized keyframe extraction method inspired by traditional graph search algorithms. \textsc{TASKER} combines task-driven relevance estimation with scene-aware temporal dynamics to select a compact yet informative set of frames that preserves essential procedural evidence while discarding redundant content. This design allows \textsc{TASKER} to operate effectively across both VideoQA and video-guided agentic tasks, providing a unified mechanism for temporal information selection.

Extensive experiments demonstrate that \textsc{TASKER} achieves both high accuracy and strong frame efficiency across VideoQA and video-guided agentic task benchmarks, consistently outperforming prior temporal selection and video-agent methods such as VideoTree and VideoAgent. In our experiments, \textsc{TASKER} achieves 63.1\% accuracy on EgoSchema fullset~\cite{mangalam2023egoschemadiagnosticbenchmarklongform} (surpassing the best baseline by 2.0\%) and 77.4\% average accuracy on NExT-QA~\cite{xiao2021nextqanextphasequestionansweringexplaining} (surpassing the best baseline by 1.8\%).
In Figure \ref{fig:frame_efficiency}, we compare the frame efficiency of \textsc{TASKER} with two baselines, highlighting its ability to effectively identify key information. 
These results highlight the effectiveness of generalized keyframe extraction as a mechanism for bridging VideoQA and video-guided agentic tasks.

Our contributions can be summarized as follows:

\begin{itemize}
\item We identify a key limitation of existing video benchmarks and propose a two-level taxonomy that connects VideoQA with video-guided agentic tasks, highlighting the role of video in-context learning.
\item We introduce \textsc{VG-GUI-Bench}, a new benchmark that pairs video tutorials with GUI agent tasks to evaluate procedural knowledge transfer from videos.
\item We propose \textsc{TASKER}, a task-driven and scene-aware keyframe extraction algorithm that improves both accuracy and frame efficiency across VideoQA and video-guided agentic tasks.
\end{itemize}

\section{Related Work}
\label{sec:related_work}

\textbf{Video Question Answering}~~VideoQA is a core subtask of video understanding that evaluates a model's ability to reason over temporal visual content in response to language queries~\cite{xiao2021nextqanextphasequestionansweringexplaining,zhong2022videoquestionansweringdatasets,nguyen2024videolanguageunderstandingsurveymodel}. 
Recent benchmarks increasingly emphasize long-form videos and complex reasoning~\cite{mangalam2023egoschemadiagnosticbenchmarklongform,fu2024videommefirstevercomprehensiveevaluation,wu2024longvideobenchbenchmarklongcontextinterleaved,zhou2025mlvubenchmarkingmultitasklong, videobench, darkvision, rbench}. 
Early approaches relied on CNN-based visual encoders and lightweight fusion modules~\cite{he2015deepresiduallearningimage,tran2015learningspatiotemporalfeatures3d,carreira2018quovadisactionrecognition,hara2018spatiotemporal3dcnnsretrace, cvmjvideodemoireing, cvmjvideostable}. 
With the emergence of LLMs, recent methods typically integrate pretrained visual encoders with projection layers and large language models for reasoning and generation~\cite{zhang2023videollamainstructiontunedaudiovisuallanguage,wang2022internvideogeneralvideofoundation,lin2024videollavalearningunitedvisual,li2025temporalpreferenceoptimizationlongform,shen2024longvuspatiotemporaladaptivecompression,weng2024longvlmefficientlongvideo,song2024moviechatdensetokensparse, zhang2025bee}. 
Beyond end-to-end Video-LLMs, agent-based frameworks~\cite{xiao2024videoqaerallmsempirical,tang2024videounderstandinglargelanguage} further enhance reasoning through prompting, memory, tool use, and planning~\cite{zhang2024simplellmframeworklongrange,fan2024videoagentmemoryaugmentedmultimodalagent,gupta2022visualprogrammingcompositionalvisual,jeoung2024adaptivevideounderstandingagent, DBLP:conf/nips/FanCGY25, fan2025agentickeyframesearchvideo}.

\smallskip
\noindent
\textbf{Keyframe Extraction}~~Efficiently identifying informative frames is crucial to understanding long videos. 
Existing methods adopt diverse strategies, including learned frame selection~\cite{park2024framesusefulefficientstrategies}, attention-based segment extraction~\cite{yang2024doraemongptunderstandingdynamicscenes}, uniform sampling with image-grid modeling~\cite{ye2025re}, and recursive agent-guided selection~\cite{wang2024videoagentlongformvideounderstanding}. 
VideoTree~\cite{wang2024videotreeadaptivetreebasedvideo} constructs a static hierarchical tree via feature clustering and performs LLM-guided search over the tree. 
\textbf{Compared with VideoTree's precomputing features for all frames, we selectively extract information during the search process, enabling question-aware tree construction with reduced computational overhead. Our primary difference from VideoAgent~\cite{wang2024videoagentlongformvideounderstanding} lies in that our frame selection strategy explicitly accounts for intra-video scene transitions as well as the intrinsic structural organization of the video.}

\smallskip
\noindent
\textbf{Video-Guided Tasks}~~
Video-guided tasks require models to acquire a task or skill by reasoning over instructional videos~\cite{wang2025videochata1thinkinglongvideos,yu2025llmguidedscenariobasedguitesting,hu2025coschainofshotpromptinglong,yang2023setofmarkpromptingunleashesextraordinary,hu2025showuipiflowbasedgenerativemodels}. For example, Mobile-Agent-V enables agents to acquire GUI operation knowledge from instructional videos and apply it to new tasks~\cite{wang2025mobileagentvvideoguidedapproacheffortless}.
On the data and evaluation side, the community has contributed a range of datasets~\cite{jang2025scalablevideotodatasetgenerationcrossplatform,lu2025videoagenttrekcomputerusepretraining,sun2025guixploreempoweringgeneralizablegui,liu2025learnactfewshotmobilegui,zhang2026showuialohahumantaughtguiagent} and benchmarks~\cite{lin2024videoguibenchmarkguiautomation,lin2025computeruseagentsjudgesgenerative,dong2026demoiclincontextlearningprocedural,dou2026clbenchbenchmarkcontextlearning,lin2026switchbenchmarkingmodelinghandling}. 
\textbf{Compared with TongUI~\cite{tongui} and Watch-and-Learn~\cite{watch-and-learn}, which focus on converting videos to learnable trajectories, our TASKER method works at the frame selection level as a training-free module for both VideoQA and downstream agentic tasks.}


\section{Method}
\label{sec:method}

In the method section, we first present the details of \textsc{VG-GUI-Bench}, followed by the \textsc{TASKER} algorithm. The preliminaries on classical graph search algorithms are provided in \textbf{Appendix A}.

\subsection{\textsc{VG-GUI-Bench}}

Despite growing interest in video-guided GUI agents, there remains a lack of high-quality benchmarks for evaluating MLLMs on long-horizon GUI task execution from video tutorials. To fill this gap, we construct \textsc{VG-GUI-Bench} (Video-Guided GUI Benchmark), a dedicated benchmark for systematically assessing MLLM performance on video-guided, long-horizon GUI tasks.

\smallskip
\noindent
\textbf{Data Source}~~We build upon the high-quality dataset provided by MONDAY~\cite{jang2025scalablevideotodatasetgenerationcrossplatform}, from which we obtain input tutorial videos, ground-truth action sequences, and keyframe screenshots as evaluation references. We further design task-specific prompts to guide the model in generating predicted actions at each step. On average, each episode contains 10.71 steps, making \textsc{VG-GUI-Bench} a genuinely long-horizon benchmark that requires sustained reasoning over extended action sequences.
In total, \textsc{VG-GUI-Bench} has 1,000 test cases.

\begin{figure*}[t]
  \centering
  \includegraphics[width=\linewidth]{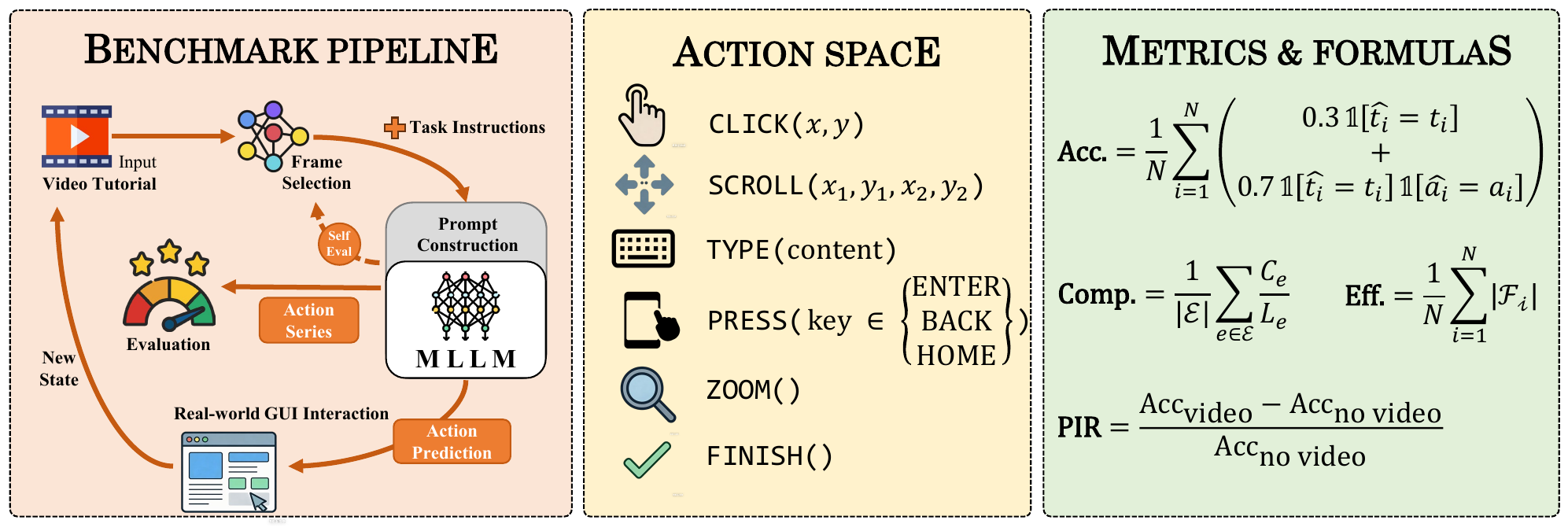}
  \caption{Overview of the \textsc{VG-GUI-Bench} benchmark, including benchmark pipeline, action space, metrics and formulas.}
  \label{fig:benchmark}
\end{figure*}

\smallskip
\noindent
\textbf{Action Space}~~Previous works often adopt inconsistent and ad-hoc action naming conventions, lacking a unified standard and a scientifically grounded taxonomy. To address this, we define a standardized action space comprising the following six action types:

\begin{itemize}
    \item \texttt{CLICK(x, y)}: Perform a tap at the coordinate $(x, y)$.
    \item \texttt{SCROLL(x1, y1, x2, y2)}: Perform a swipe or drag gesture from $(x_1, y_1)$ to $(x_2, y_2)$, covering directional interactions that require two coordinate pairs.
    \item \texttt{TYPE(content)}: Input a text string specified by the \texttt{content} argument.
    \item \texttt{PRESS(key)}: Press a system-level key, where \texttt{key} $\in$ \{\texttt{BACK}, \texttt{HOME}, \texttt{ENTER}\}. These keys carry distinct semantic meanings on mobile platforms and cannot be substituted by other operations.
    \item \texttt{ZOOM()}: Perform a pinch-to-zoom gesture. This action takes no arguments.
    \item \texttt{FINISH()}: Indicate that the task has been completed. This action takes no arguments.
\end{itemize}

\smallskip
\noindent
\textbf{Evaluation Pipeline}~~As illustrated in Figure~\ref{fig:benchmark}, the evaluation pipeline of \textsc{VG-GUI-Bench} operates as follows. The input tutorial video is first processed by a frame selection module, which extracts relevant keyframes from the video. The selected frames, together with the task instruction, are then used to construct a prompt that is fed into the MLLM to predict the next GUI action. The predicted action is executed in the GUI environment, producing a new interaction state. Based on this state, the model continues predicting subsequent actions step by step. The predicted actions are finally evaluated against the ground-truth action sequence by \textsc{VG-GUI-Bench}. This process repeats iteratively until the episode terminates.

\smallskip
\noindent
\textbf{Evaluation Metrics}~~We propose four complementary metrics to provide a comprehensive assessment:

\begin{itemize}

\item \textbf{Accuracy} measures the correctness of individual action predictions. 
A prediction receives a score of $0.3$ if the action type is correct, and an additional $0.7$ if the action arguments are also correct. 
For argument-free actions (\texttt{ZOOM}, \texttt{FINISH}), a correct type prediction yields the full score.
In addition to the overall accuracy, we also report the type accuracy and score breakdown for each individual action category to provide finer-grained insights into model behavior.

\begin{equation}
\text{Acc.} =
\frac{1}{N}
\sum_{i=1}^{N}
\mathbbm{1}\left(\hat{t}_i = t_i\right)
\left(
0.3 + 0.7\times\mathbbm{1}\left(\hat{a}_i = a_i\right)
\right)
\end{equation}

where $N$ denotes the total number of steps, $t_i$ and $\hat{t}_i$ are the ground-truth and predicted action types, and $a_i$ and $\hat{a}_i$ are the ground-truth and predicted action arguments.

\item \textbf{Completion} measures the proportion of correctly executed steps within each episode, averaged across all episodes:

\begin{equation}
\text{Comp.} =
\frac{1}{|\mathcal{E}|}
\sum_{e \in \mathcal{E}}
\frac{C_e}{L_e}
\end{equation}

where $\mathcal{E}$ denotes the set of all episodes, $C_e$ is the number of correctly executed steps in episode $e$, and $L_e$ is the total number of steps in episode $e$.

\item \textbf{Efficiency} measures the average number of input frames consumed per prediction step, reflecting the computational cost of the frame selection strategy:

\begin{equation}
\text{Eff.} =
\frac{1}{N}
\sum_{i=1}^{N}
|\mathcal{F}_i|
\end{equation}

where $|\mathcal{F}_i|$ denotes the number of frames provided to the MLLM at step $i$.

\item \textbf{Performance Improvement Rate (PIR)} 
measures how much the model can learn from the video:

\begin{equation}
\text{PIR} =
\frac{\text{Acc}_{\text{video}} - \text{Acc}_{\text{no video}}}
{\text{Acc}_{\text{no video}}}
\end{equation}

where $\text{Acc}_{\text{no video}}$ and $\text{Acc}_{\text{video}}$ denote the accuracy without and with video tutorial input respectively.

\end{itemize}

\subsection{\textsc{TASKER} Algorithm}

\begin{figure*}[t]
  \centering
   \includegraphics[width=\linewidth]{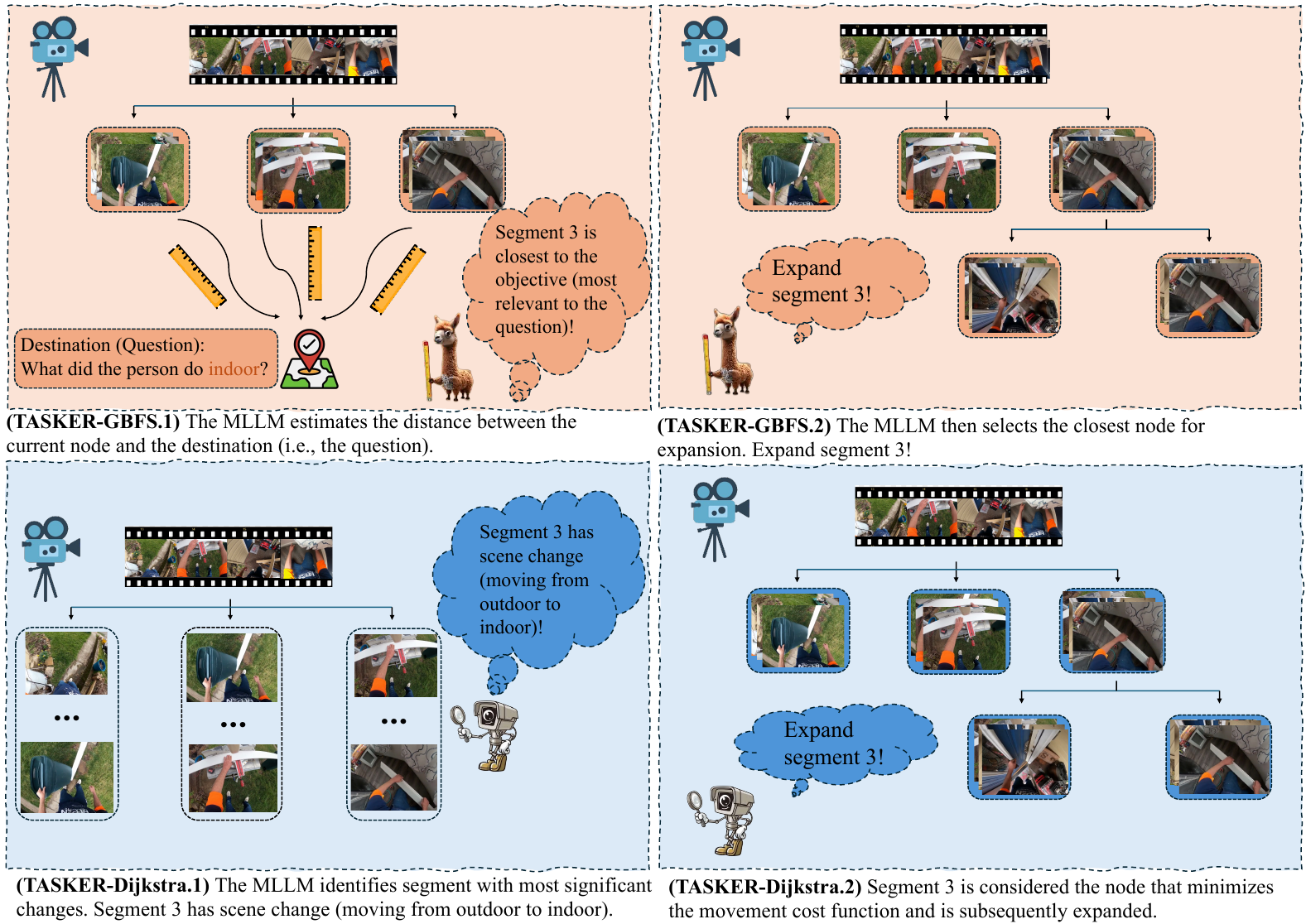}
    \caption{Illustration of \textsc{TASKER}'s cost function evaluation and node expansion steps. \textsc{TASKER-GBFS} variant evaluates distance based on question relevance. \textsc{TASKER-Dijkstra} variant evaluates distance based on scene dynamics.}
   \label{fig:grid}
\end{figure*}

In this section, we first define the search objective, nodes, cost function, and termination conditions in our \textsc{TASKER} algorithm, providing a comprehensive overview of the tree-structured keyframe search process. We also explain how the algorithm utilizes the retrieved information to answer questions. The key steps of \textsc{TASKER} (leveraging MLLMs to evaluate the cost function and node expansion) are illustrated in Figure \ref{fig:grid}.

\noindent
\textbf{Search Objective}~~
The search objective is to identify a sufficient set of keyframes whose combined information is sufficient to answer the question or complete the agentic task. The ultimate goal is to answer the question. 
\\
\noindent
\textbf{Nodes Expansion}~~In \textsc{TASKER} algorithm, we divide the video into multiple video segments, with each video segment representing a node. The initial node $\mathcal{N}_0$ is the entire video, which is first uniformly split into $M$ segments, where $M$ is a tunable hyper-parameter. These video segments are then put into an open list $\mathcal{L}$. The next node to be expanded, or the next video segment to be processed, is selected based on the cost function $f(n)$ we define. The expansion process means further subdividing the selected video segment. In this work, we perform a binary split on the segment for node expansion.\\
\noindent
\textbf{Answer Prediction}~~We define the first and last frames of all current video segments as \textbf{Visible Frames} $\mathcal{F}_{v}$. They are connected to each other, meaning that the last frame of one video segment is the first frame of the next. We can fully utilize the information in the visible frames, while the information in the other frames is temporarily inaccessible. For the visible frames, we can employ two approaches: directly inputting the frames into the MLLM or first generating captions and then performing reasoning in the textual modality. In either way, we predict an answer or action based on the information from the visible frames. 
\noindent
\textbf{Cost Function}~~Leveraging the evaluation capability of the MLLMs, we design different cost functions based on different basic search algorithms.
\begin{itemize}
    \item \textbf{\textsc{TASKER-GBFS}: Task-Driven Cost Function}~~In Greedy Best-First Search (GBFS), the cost function $h(n)$ represents the distance from the current node to the destination. Accordingly, we let the MLLM evaluate the current visible frame's information and identify what visual information is missing for answering the question. The missing information can be seen as the distance between the current node and the destination. GBFS algorithm selects the node with the smallest $h(n)$ to expand. In our adaptation, the MLLM attempts to identify where the missing visual information is likely located between which two specific invisible frames, determining which video segments should be expanded. This leads to a variant of the \textsc{TASKER} algorithm named \textsc{TASKER-GBFS}.

    \item \textbf{\textsc{TASKER-Dijkstra}: Scene-Aware Cost Function}~~
    In Dijkstra's algorithm, the cost function $g(n)$ represents the cost of reaching the current node from the start point. In our adaptation, the MLLM evaluates the visible frames to identify the video segment with the most significant scene change, such as transitions in scenes, figures, or activities between the first and last frames of a segment. 
    Segments with larger scene changes are considered more informative and thus prioritized for further exploration, corresponding to nodes with smaller $g(n)$. 
    Notably, in classical Dijkstra's algorithm, the cost function does not depend on the destination; similarly, in \textsc{TASKER-Dijkstra}, the MLLM is not driven by the task or question, but instead selects keyframes based on scene changes and the intrinsic structure of the video, making the search purely scene-aware.
    \item \textbf{\textsc{TASKER-A*}: Task Driven \& Scene Aware Cost Function}~~For the A* Algorithm, the cost function is the sum of the heuristic evaluation function and the movement cost function, i.e., $f(n) = h(n) + g(n)$, which means that A* Algorithm takes into account both the distance from the current node to the destination and the distance from the start point to the current node. Correspondingly, in our \textsc{TASKER-A*} variant, the MLLM must simultaneously consider two factors: (1) which video segment is likely to contain the missing information, and (2) which video segment exhibits the most significant scene change. Only video segments that satisfy both are prioritized for expansion.
    \item \textbf{\textsc{TASKER-BFS}: Naive Cost Function}~~We also propose a naive algorithm variant, \textsc{TASKER-BFS} which does not rely on a MLLM to evaluate the cost function. Instead, it performs a breadth-first expansion, continually splitting all the existing video segments (in the case of no pruning). Like BFS, \textsc{TASKER-BFS} advances in a wave-like manner, steadily progressing. This variant is suitable for situations where a MLLM cannot be accessed, or where the overhead introduced by the MLLM is less of a concern, with a greater emphasis on ensuring no information is overlooked.
\end{itemize}
We do not intend to introduce \textsc{TASKER-DFS} variant, as its depth-first expansion focuses on a single initial segment. Without strict termination conditions, it is prone to falling into local optima.\smallskip\\
\textbf{Termination Condition and Confidence Estimation}~~
Traditional search algorithms rely on deterministic termination conditions, typically defined by whether the search objective has been reached. In contrast, for keyframe search in VideoQA and Video-Guided Agentic Tasks, such conditions are difficult to define, as it is unclear whether sufficient information has been collected. To address this issue, we leverage the reflection and self-evaluation capabilities of MLLMs to estimate the confidence of the predicted answer and determine whether to terminate the search. \textsc{TASKER} stops when the model produces a sufficiently confident prediction. Specifically, we combine two confidence estimation methods through a voting mechanism, as described below.

\begin{algorithm}[t]
\caption{Task-driven And Scene-aware Keyframe searchER (\textsc{TASKER})}
\label{algorithm_1}
\begin{algorithmic}[1]
\Require Video $v$, question $q$, MLLM $F$, confidence threshold $C$, max iteration $T$, uniform sampling size $M$, beam size $B$, frozen frame set $\mathcal{S}_{\text{frozen}}$
\Ensure Answer $\hat{y}$, keyframes $\{\mathcal{F}_k\}$
\State $\mathcal{N}_0 \gets v$
\State $\mathcal{N}_1, \mathcal{N}_2, ... ,\mathcal{N}_{M}\gets \text{UniformSegment}(\mathcal{N}_0, M)$
\State $\mathcal{L} \gets \{\mathcal{N}_0, \mathcal{N}_1, ... , \mathcal{N}_M\}$
\State $\mathcal{S}_{\text{frozen}} \gets \emptyset$
\State $t \gets 1$
\While{$t \leq T$}
    \State $\{\mathcal{F}_v\} \gets \text{ExtractVisibleFrames}(\mathcal{L})$
    \State $\hat{y} \gets \text{PredictAnswer}(F, \{\mathcal{F}_v\}, q)$
    \State $c_1, c_2 \gets \text{EvaluateConfidence}(F, \hat{y}, \{\mathcal{F}_v\}, q)$
    \If{$c_1 \geq C\, \text{and}\, c_2 \geq C$}
        \State \textbf{break} 
    \Else
        \State $\{p\} \gets \text{EvaluateCostFunction}(F, \{\mathcal{F}_v\}, \mathcal{L} \setminus \mathcal{S}_{\text{frozen}})$
        \State $\mathcal{L} \gets \text{SelectAndExpandNodes}(\mathcal{L}, \{p\}, B)$
        \State $\mathcal{L}, \mathcal{S}_{\text{frozen}} \gets \text{ValidateNewFrame}(\mathcal{L}, \mathcal{S}_{\text{frozen}}, F, q)$
    \EndIf
    \State $t \gets t+1$
\EndWhile
\State $\{\mathcal{F}_k\} \gets \{\mathcal{F}_v\}$
\State \Return $\hat{y}$, $\{\mathcal{F}_k\}$
\end{algorithmic}
\end{algorithm}

\begin{itemize}
    \item \textbf{Self-Evaluation and Self-Reflection}~~
    MLLMs can be instructed to self-evaluate their responses, reflecting on potential shortcomings in their responses~\cite{shinn2023reflexionlanguageagentsverbal, ren2023selfevaluationimprovesselectivegeneration}. Therefore, after generating an answer, we input the question, information of visible frames, and the MLLM's previous reasoning chain and predicted answer back into the model. 
    The MLLM then assesses the accuracy and reliability of its previous answer and output a confidence score ($c_1$).
    
    \item 
    \textbf{Temporal Summarization}~~
    The captions of the sampled frames are discrete.
    To integrate the sampled frames along the temporal dimension, we instruct the MLLM to summarize their captions to form a cohesive overview of the video.
    We use few-shot examples~\cite{brown2020languagemodelsfewshotlearners} to generate a more accurate and detailed video summary. Then we prompt the MLLM to predict the answer and output a confidence score ($c_2$) based on the summary. The advantage of this approach is to consider the sampled frames in a complete temporal context rather than in isolation.
\end{itemize}
We employ a voting mechanism to ensemble the above two methods. The search process only terminates when both methods independently determine that they have sufficient confidence ($c_1 \geq C\, \text{and}\, c_2 \geq C$, $C$ is the threshold).

Additionally, after each expansion, we apply a \textbf{frame validation} step: each newly revealed frame is first checked for visual redundancy against existing visible frames, and then assessed by the MLLM for task relevance. Redundant or irrelevant frames are discarded (with nearby relevant replacements searched when possible), and segments that yield only redundant frames are added to a frozen set $\mathcal{S}_{\text{frozen}}$ to avoid repeated exploration.
We summarize \textsc{TASKER} algorithm in Algorithm \ref{algorithm_1}. 


\section{Experiments}
\label{sec:experiments}
\begin{table}[ht!]
\caption{\textbf{Comparison between \textsc{TASKER} and other methods.} We highlight the gain of our method over VideoTree~\cite{wang2024videotreeadaptivetreebasedvideo} in blue.}
\centering
\setlength{\tabcolsep}{1.4mm}
\resizebox{\linewidth}{!}{%
\begin{tabular}{lccccccc}
\toprule
\textbf{Model} & \textbf{(M)LLM} & \multicolumn{2}{c}{\textbf{EgoSchema}} & \multicolumn{4}{c}{\textbf{NExT-QA}} \\
\cmidrule(lr){3-4} \cmidrule(lr){5-8}
 & & \textbf{Sub.} & \textbf{Full} & \textbf{Tem.} & \textbf{Cau.} & \textbf{Des.} & \textbf{Avg.} \\
\midrule
\multicolumn{8}{l}{\textbf{Based on Open-source Captioners and Open-source LLMs}} \\
\midrule
MVU~\cite{ranasinghe2025understandinglongvideosmultimodal}& Mistral-13B & 60.3 & 37.6 & 55.4 & 48.1 & 64.1 & 55.2 \\
LangRepo~\cite{kahatapitiya2024languagerepositorylongvideo} & Mixtral-8x7B & 66.2 & 41.2 & 51.4 & 64.4 & 69.1 & 60.9 \\
Video-LLA+INTP~\cite{shang2024interpolatingvideollmslongersequencelmms} & Vicuna-7B v1.5 & - & 38.6 & 58.6 & 61.9 & 72.2 & 62.7 \\
\midrule
\multicolumn{8}{l}{\textbf{Based on Proprietary MLLMs}} \\
\midrule
IG-VLM~\cite{kim2024imagegridworthvideo} & GPT-4V & 59.8 & - & 63.6 & 69.8 & 74.7 & 68.6 \\
LVNet~\cite{park2024framesusefulefficientstrategies} & GPT-4o & 68.2 & 61.1 & 65.5 & 75.0 & 81.5 & 72.9 \\
\midrule
\multicolumn{8}{l}{\textbf{Based on Open-source Captioners and Proprietary LLMs}} \\
\midrule
ProViQ~\cite{choudhury2023zeroshotvideoquestionanswering} & GPT-3.5 & 57.1 & - & - & - & - & 64.6 \\
MoReVQA~\cite{min2024morevqaexploringmodularreasoning}& PaLM-2 & - & 51.7 & 64.6 & 70.2 & - & 69.2 \\
Vamos~\cite{wang2024vamosversatileactionmodels} & GPT-4 & 51.2 & 48.3 & - & - & - & - \\
LLoVi~\cite{zhang2024simplellmframeworklongrange} & GPT-4 & 61.2 & - & 61.0 & 69.5 & 75.6 & 67.7 \\
VideoAgent~\cite{wang2024videoagentlongformvideounderstanding} & GPT-4 & 60.2 & 54.1 & 64.5 & 72.7 & 81.1 & 71.3 \\
VideoAgent~\cite{fan2024videoagentmemoryaugmentedmultimodalagent} & GPT-4 & 62.8 & 60.2 & - & - & - & - \\
LifelongMemory~\cite{wang2024lifelongmemoryleveragingllmsanswering} & GPT-4 & 64.1 & 58.6 & - & - & - & - \\
VideoTree~\cite{wang2024videotreeadaptivetreebasedvideo} & GPT-4 & 66.2 & 61.1 & 70.6 & 76.5 & 83.9 & 75.6 \\
\textbf{\textsc{TASKER} (Ours)} & GPT-4 & \textbf{68.0 (\textcolor{blue}{1.8 $\uparrow$})} & \textbf{63.1 (\textcolor{blue}{2.0 $\uparrow$})} & \textbf{72.3 (\textcolor{blue}{1.7 $\uparrow$})} & \textbf{78.2 (\textcolor{blue}{1.7 $\uparrow$})} & \textbf{85.4 (\textcolor{blue}{1.5 $\uparrow$})} & \textbf{77.4 (\textcolor{blue}{1.8 $\uparrow$})} \\
\textbf{\textsc{TASKER} (Ours)} & GPT-4o & \textbf{68.6 (\textcolor{blue}{2.4 $\uparrow$})} & \textbf{63.6 (\textcolor{blue}{2.5 $\uparrow$})} & \textbf{72.9 (\textcolor{blue}{2.3 $\uparrow$})} & \textbf{79.0 (\textcolor{blue}{2.5 $\uparrow$})} & \textbf{86.1 (\textcolor{blue}{2.2 $\uparrow$})} & \textbf{78.1 (\textcolor{blue}{2.5 $\uparrow$})} \\
\midrule
\multicolumn{8}{l}{\textbf{Based on Open-source MLLMs}} \\
\midrule
VideoAgent~\cite{wang2024videoagentlongformvideounderstanding} & Qwen3-VL & 75.8 & 74.6 & 79.7 & 83.5 & 86.5 & 82.8 \\
VideoTree~\cite{wang2024videotreeadaptivetreebasedvideo} & Qwen3-VL & 77.2 & 76.7 & 81.9 & 85.1 & 86.3 & 84.3 \\
\textbf{\textsc{TASKER} (ours)} & Qwen3-VL & \textbf{79.4 (\textcolor{blue}{2.2 $\uparrow$})} & \textbf{77.3 (\textcolor{blue}{0.6 $\uparrow$})} & \textbf{83.6 (\textcolor{blue}{1.7 $\uparrow$})} & \textbf{85.3 (\textcolor{blue}{0.2 $\uparrow$})} & \textbf{87.5 (\textcolor{blue}{1.2 $\uparrow$})} & \textbf{85.1 (\textcolor{blue}{0.8 $\uparrow$})} \\
\bottomrule
\end{tabular}
}
\label{table:main_results}
\end{table}

Our experiments are conducted on two VideoQA benchmarks, EgoSchema~\cite{mangalam2023egoschemadiagnosticbenchmarklongform} and NExT-QA~\cite{xiao2021nextqanextphasequestionansweringexplaining}, as well as the proposed video-guided agentic task benchmark, \textsc{VG-GUI-Bench}. Further details of EgoSchema and NExT-QA benchmarks can be found in \textbf{Appendix B}.

\subsection{Main Results}

\subsubsection{VideoQA Results}

We compare the performance of \textsc{TASKER} with various related approaches on LLM-driven VideoQA using \textsc{TASKER-A*} variant. Most of the baselines are mentioned in the related work and \textbf{Appendix A}. Implementation details are provided in \textbf{Appendix B}. Prompts we use are listed in \textbf{Appendix C}. Table \ref{table:main_results} demonstrates that \textsc{TASKER} significantly outperforms all these baselines. Specifically, \textsc{TASKER} (with GPT-4 as base LLM) achieves 63.1\% accuracy on EgoSchema fullset (surpassing the best baseline by 2.0\%) and 77.4\% accuracy on NExT-QA (surpassing the best baseline by 1.8\%). 
Results based on the open-source MLLM (Qwen3-VL-235B-A22B-Instruct) also show that TASKER consistently outperforms VideoTree~\cite{wang2024videotreeadaptivetreebasedvideo} and VideoAgent~\cite{wang2024videoagentlongformvideounderstanding}.

Moreover, \textsc{TASKER} operates in a training-free, zero-shot setting, while it still outperforms training-based methods such as LVNet~\cite{park2024framesusefulefficientstrategies} and Vamos~\cite{wang2024vamosversatileactionmodels}. Meanwhile, \textsc{TASKER} processes only visible frames, for instance, achieving the reported performance requires only about 15\% of the total frames. In contrast, methods like LangRepo~\cite{kahatapitiya2024languagerepositorylongvideo} and LifelongMemory~\cite{wang2024lifelongmemoryleveragingllmsanswering} process all frames without selection. 

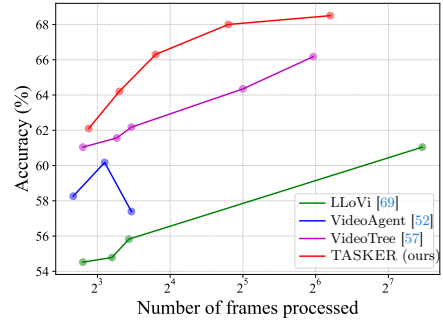
\begin{wrapfigure}{r}{0.48\textwidth}  
  \centering
  \footnotesize
  \begin{overpic}[width=\linewidth]{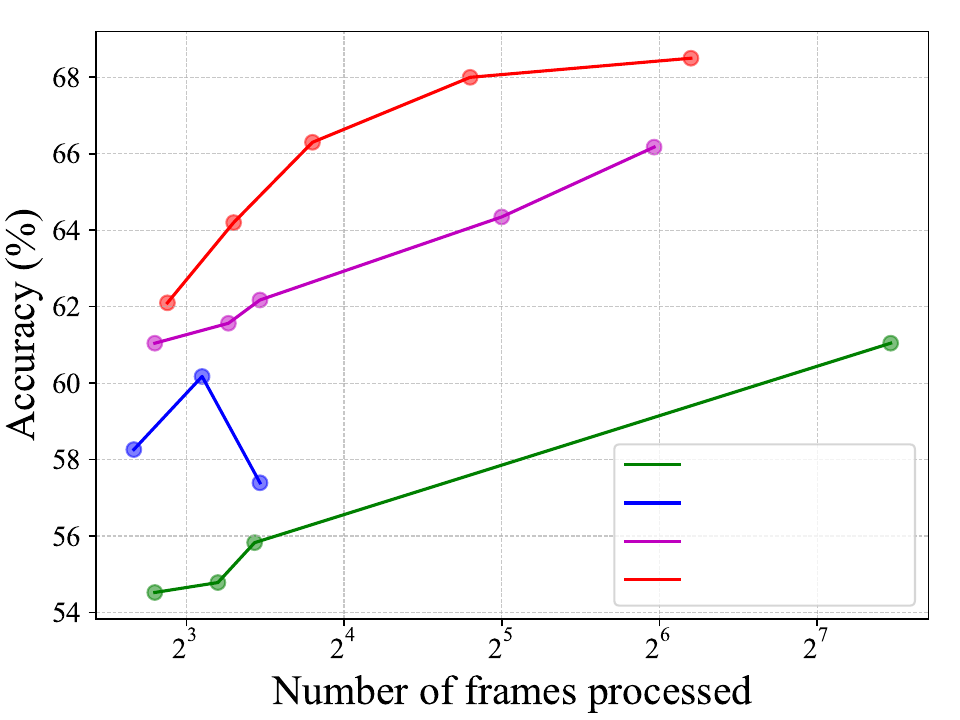}
    \put(72, 25.7){\scalebox{0.6}{LLoVi~\cite{zhang2024simplellmframeworklongrange}}}
    \put(72, 21.6){\scalebox{0.6}{VideoAgent~\cite{wang2024videoagentlongformvideounderstanding}}}
    \put(72, 17.5){\scalebox{0.6}{VideoTree~\cite{wang2024videotreeadaptivetreebasedvideo}}}
    \put(72, 13.7){\scalebox{0.6}{\textsc{TASKER} (ours)}}
  \end{overpic}
  \vspace{-10pt}
  \caption{Demonstration of \textsc{TASKER}'s high frame efficiency. When processing the same number of video frames with the same (M)LLM, \textsc{TASKER} achieves higher QA accuracy.
  }
  \label{fig:frame_efficiency}
\end{wrapfigure}

Additionally, as shown in Figure \ref{fig:frame_efficiency}, we compare \textsc{TASKER}'s frame efficiency with other keyframe extraction methods in the same condition on EgoSchema~\cite{mangalam2023egoschemadiagnosticbenchmarklongform} subset. \textbf{At the same accuracy level (66\%), \textsc{TASKER} uses only about 1/4 of the frames required by VideoTree.} Moreover, VideoTree clusters features of all frames during preprocessing, whereas \textsc{TASKER} only has access to visible frames and does not utilize information from the rest. The results show that \textsc{TASKER} utilizes frames more efficiently than LLoVi~\cite{zhang2024simplellmframeworklongrange}, VideoAgent~\cite{wang2024videoagentlongformvideounderstanding} and VideoTree~\cite{wang2024videotreeadaptivetreebasedvideo}, demonstrating its superior ability to identify key information.



\subsubsection{Video-Guided Agentic Task Results}


\begin{table}[t]
\centering
\setlength{\tabcolsep}{1.2mm}
\caption{\textbf{\textsc{VG-GUI-Bench} leaderboard.}}
\label{tab:leaderboard}
\resizebox{\columnwidth}{!}{%
\begin{tabular}{cll cccc}
\toprule
\textbf{Rank} & \textbf{Model} & \textbf{Mode} & \textbf{Acc.(\%)} & \textbf{Type Acc.(\%)} & \textbf{Comp.(\%)} & \textbf{PIR} \\
\midrule
\multirow{2}{*}{1} & \multirow{2}{*}{Gemini-3.1-Pro} & No Video & 58.51 & 74.58 & 78.53 & -- \\
 & & 10 Uniform Frames & 61.68 & 76.25 & 78.61 & 0.054 \\
\midrule
\multirow{2}{*}{2} & \multirow{2}{*}{GPT-5-mini} & No Video & 55.76 & 71.93 & 75.97 & -- \\
 & & 10 Uniform Frames & 58.40 & 75.27 & 78.96 & 0.047 \\
\midrule
\multirow{2}{*}{3} & \multirow{2}{*}{Kimi-K2.5} & No Video & 57.22 & 72.91 & 76.31 & -- \\
 & & 10 Uniform Frames & 58.22 & 74.78 & 78.72 & 0.017 \\
\midrule
\multirow{2}{*}{4} & \multirow{2}{*}{Claude-Sonnet-4.6} & No Video & 44.35 & 67.81 & 72.24 & -- \\
 & & 10 Uniform Frames & 45.80 & 67.71 & 72.35 & 0.033 \\
\midrule
\multirow{2}{*}{5} & \multirow{2}{*}{Seed-2.0-Pro} & No Video & 35.93 & 70.66 & 74.75 & -- \\
 & & 10 Uniform Frames & 39.78 & 75.96 & 79.36 & 0.107 \\
\midrule
\multirow{2}{*}{6} & \multirow{2}{*}{Qwen3-VL-235B-A22B} & No Video & 25.90 & 66.63 & 69.73 & -- \\
 & & 10 Uniform Frames & 26.88 & 67.42 & 71.69 & 0.038 \\
\midrule
\multirow{2}{*}{7} & \multirow{2}{*}{Gemini-3.1-Flash} & No Video & 24.73 & 67.03 & 70.54 & -- \\
 & & 10 Uniform Frames & 26.02 & 69.19 & 72.60 & 0.052 \\
\bottomrule
\end{tabular}
}
\end{table}

\begin{table*}[ht!]

\caption{\textbf{\textsc{TAKSER} and Baselines' Results on \textsc{VG-GUI-Bench}.} We report Acc., Type Acc., per-action scores, Comp., Eff., and PIR. For the keyframe selection methods, best and second-best results are in red and blue. For all methods, we use Qwen3-VL-235B-A22B-Instruct~\cite{bai2025qwen3vltechnicalreport} as the base LLM.}
\centering
\setlength{\tabcolsep}{1.2mm}
\resizebox{\linewidth}{!}{%
\begin{tabular}{lcc cccccc ccc}
\toprule
\multirow{2}{*}{\textbf{Method}} & \multicolumn{2}{c}{\textbf{Overall(\%)}} & \multicolumn{6}{c}{\textbf{Per-Action Score(\%)}} & \multirow{2}{*}{\textbf{Comp.(\%)}} & \multirow{2}{*}{\textbf{Eff. $\downarrow$}} & \multirow{2}{*}{\textbf{PIR}} \\
\cmidrule(lr){2-3} \cmidrule(lr){4-9}
 & \textbf{Acc.} & \textbf{Type Acc.} & \textbf{CLICK} & \textbf{SCROLL} & \textbf{TYPE} & \textbf{PRESS} & \textbf{ZOOM} & \textbf{FINISH} & & & \\
\midrule
\multicolumn{12}{l}{\textit{Reference Baselines}} \\
\midrule
No Video & 25.32 & 65.85 & 26.89 & 26.18 & 3.66 & 29.23 & 0 & 0 & 69.03 & 0 & - \\
All Keyframes & 37.21 & 65.75 & 49.48 & 2.38 & 25.81 & 5.26 & 0 & 0 & 72.01 & 13.23 & 0.470 \\
Uniform Sampling & 39.82 & 66.34 & 52.90 & 5.25 & 15.71 & 6.50 & 0 & 0 & 70.64 & 10.88 & 0.573 \\
Oracle Keyframes & 44.32 & 73.31 & 60.34 & 1.92 & 0 & 0 & 0 & 0 & 76.32 & 1 & 0.750 \\

\midrule
\multicolumn{12}{l}{\textit{Keyframe Selection Methods}} \\
\midrule
VideoTree~\cite{wang2024videotreeadaptivetreebasedvideo} & \textcolor{blue}{40.79} & 67.52 & \textcolor{blue}{53.41} & 7.83 & 14.75 & \textcolor{red}{6.50} & 0 & 0 & 71.93 & 10.00 & \textcolor{blue}{0.611} \\
VideoAgent~\cite{wang2024videoagentlongformvideounderstanding} & 39.86 & 67.03 & 51.61 & \textcolor{red}{10.70} & 13.81 & 5.26 & 0 & 0 & 71.17 & \textcolor{red}{5.12} & 0.574 \\
\textsc{TASKER-BFS} & 40.01 & \textcolor{blue}{69.97} & 51.87 & 9.66 & \textcolor{blue}{17.16} & 5.26 & 0 & 0 & \textcolor{blue}{73.77} & 6.10 & 0.580 \\
\textsc{TASKER-GBFS} & 40.26 & \textcolor{blue}{69.97} & 52.10 & \textcolor{blue}{9.98} & 16.75 & 5.26 & 0 & 0 & 73.75 & \textcolor{blue}{5.81} & 0.590 \\
\textsc{TASKER-Dijkstra} & 40.75 & \textcolor{red}{71.05} & 52.84 & 9.66 & \textcolor{red}{17.48} & 5.26 & 0 & 0 & \textcolor{red}{74.39} & 5.88 & 0.609 \\
\textsc{TASKER-A*} & \textcolor{red}{40.96} & 67.71 & \textcolor{red}{53.68} & 8.71 & 13.81 & \textcolor{red}{6.50} & 0 & 0 & 71.38 & 8.24 & \textcolor{red}{0.618} \\
\bottomrule
\end{tabular}
}
\label{tab:vg_gui_bench}
\end{table*}

In Table \ref{tab:leaderboard}, we provide a \textsc{VG-GUI-bench} leaderboard with 7 frontier models. Gemini-3.1-Pro consistently achieves the best overall accuracy, regardless of whether no video input is used or 10 frames are uniformly sampled from the video. Across most models, incorporating 10 uniformly sampled frames consistently improves performance over the no-video setting, demonstrating the benefit of temporal visual information for GUI understanding. Notably, Seed-2.0-Pro exhibits the largest gain in overall accuracy, improving from 35.93\% to 39.78\%.

Our \textsc{TASKER} method's results on \textsc{VG-GUI-Bench} are presented in Table \ref{tab:vg_gui_bench}. We consider the following four baselines as references:

\begin{itemize}
    \item \textbf{No Video}: The model performs the GUI agent task without taking the video tutorial as input. Since it cannot refer to the tutorial, this baseline yields the weakest performance.
    
    \item \textbf{All Keyframes}: We provide the set of tutorial frames corresponding to all moments where action behaviors occur. With access to these references, the model can closely follow the demonstrated procedure and achieves relatively strong results.
    
    \item \textbf{Uniform Sampling}: We uniformly sample frames from the full video. Although this does not guarantee coverage of key instructional moments, the stable global context allows it to outperform the \textit{All Keyframes} baseline.
    
    \item \textbf{Oracle Keyframe}: We provide the specific tutorial frame corresponding to the current GUI action with Set-of-Mark~\cite{yang2023setofmarkpromptingunleashesextraordinary} annotations, allowing the model to ``peek at the answer''. While this enables high scores by copying visual targets, it encourages over-reliance on visual imitation, causing complete failure on operations like \texttt{TYPE} and \texttt{PRESS}.

\end{itemize}

We also compare against representative keyframe selection methods, VideoTree and VideoAgent. Overall, our \textsc{TASKER} framework demonstrates excellent performance. Specifically, \textsc{TASKER-A*} achieves the highest Overall Acc.\ (40.96), PIR (0.618), and \texttt{CLICK} score (53.68), outperforming the strong VideoTree baseline. Furthermore, compared to other dynamic selection methods, \textsc{TASKER} effectively deduplicates redundant frames, yielding superior accuracy with higher efficiency (fewer frames per step). Beyond accuracy, our approach attains high task completion rates, with \textsc{TASKER-Dijkstra} (74.39) closely approaching the \textit{Oracle Keyframe} upper bound. Finally, the varied search strategies demonstrate the framework's internal diversity, robustness, and strong extensibility.

In addition to the VG-GUI-Bench benchmark, we also collected accompanying videos for OSWorld~\cite{xie2024osworld} and evaluated whether the \textsc{TASKER} method can improve the capabilities of GUI agents on OSWorld. Detailed results are provided in \textbf{Appendix G}.

\subsection{Ablation Studies}

\subsubsection{Basic Search Algorithms}

In Table \ref{tab:vg_gui_bench}, we have already compared various variants of the \textsc{TASKER} algorithm and observed that \textsc{TASKER-A*} achieves the best performance.

\begin{wraptable}[10]{r}{0.50\columnwidth} 
\vspace{-30pt}
\caption{\textbf{Ablation on basic search algorithms}. We highlight the improvement of \textsc{TASKER-A*} over the naive \textsc{TASKER-BFS}, emphasizing the role of cost-function evaluation.}
\label{tab:search_algorithm}
\centering
\small
\setlength{\tabcolsep}{1.6mm}
\renewcommand{\arraystretch}{0.95}
\resizebox{\linewidth}{!}{%
\begin{tabular}{@{}lcc@{}}
\toprule
\textbf{Algorithm} & \textbf{Accuracy} & \textbf{\# Visible} \\
\midrule
\textsc{TASKER-BFS}      & 64.7 & 31.2 \\
\textsc{TASKER-GBFS}     & 67.0 & \textbf{27.3} \\
\textsc{TASKER-Dijkstra} & 66.8 & 27.6 \\
\textsc{TASKER-A*} & \textbf{68.0 (\textcolor{blue}{3.3 $\uparrow$})} & 27.9 \\
\bottomrule
\end{tabular}
}
\end{wraptable}

In Table \ref{tab:search_algorithm}, we further investigate performance and frame efficiency of several \textsc{TASKER} algorithm variants with different base search algorithms on the EgoSchema subset. The frame efficiency is measured by the number of visible frames. Specifically, on EgoSchema dataset, all videos are three minutes long, and we set the initial frame rate $\text{fps}=1$, which means the initial overall frame number is 180. We observe that \textsc{TASKER-A*} achieves the highest accuracy. \textsc{TASKER-BFS} ranks last in accuracy and has the lowest frame efficiency, as \textsc{TASKER-BFS} algorithm exhaustively explores all branches, leading to higher exploration costs. \textsc{TASKER-GBFS} slightly outperforms \textsc{TASKER-Dijkstra} on both metrics, while \textsc{TASKER-A*} combines the strengths of them, significantly improving accuracy with only a slight compromise in frame efficiency. This demonstrates that efficient keyframe localization requires both the heuristic search function and the movement cost function.

\subsubsection{Termination Condition}

In Table \ref{tab:termination_condition}, we conduct ablation experiments on the termination condition of the search process, using \textsc{TASKER-A*} on the EgoSchema subset. 


\begin{wraptable}[10]{r}{0.47\columnwidth}
\vspace{-30pt}
\hspace{-2mm}
\caption{\textbf{Ablation on termination condition}}
\label{tab:termination_condition}
\centering
\small
\setlength{\tabcolsep}{0.9mm}
\renewcommand{\arraystretch}{0.95}

\resizebox{\linewidth}{!}{%
\begin{tabular}{@{}lcc@{}}
\toprule
\textbf{Method} & \textbf{Accuracy} & \textbf{\# Visible Frames}\\
\midrule
Self-Evaluation & 67.4 & \textbf{27.4} \\
Summarization & 67.3 & 28.2 \\
Vote & \textbf{68.0} & 27.9 \\
\bottomrule
\end{tabular}%
}
\end{wraptable}

The results show that self-evaluation \& self-reflection and temporal summarization assess information sufficiency from different perspectives. When combined, they enhance the reliability of confidence estimation, leading to improved algorithm performance.

\subsubsection{Base LLM}

We also conducted an ablation study on the base LLM of the \textsc{TASKER} algorithm in Table \ref{tab:base_llm}. 


\begin{wraptable}[5]{r}{0.47\columnwidth}
\vspace{-70pt}
\hspace{-2mm}
\caption{\textbf{Ablation on different base LLMs}}
\label{tab:base_llm}
\centering
\small
\setlength{\tabcolsep}{2.0mm}
\renewcommand{\arraystretch}{0.95}

\resizebox{\linewidth}{!}{%
\begin{tabular}{@{}lcc@{}}
\toprule
\textbf{Base LLM} & \textbf{Accuracy} & \textbf{\# Visible Frames}\\
\midrule
\textsc{GPT-4} & 68.0 & 27.9 \\
\textsc{GPT-4o} & \textbf{68.6} & \textbf{26.7} \\
\textsc{o3-mini} & 67.3 & 28.3 \\
\textsc{Deepseek-R1} & 67.6 & 26.9 \\
\textsc{LLaMA-3.3-70B} & 65.2 & 27.4 \\
\bottomrule
\end{tabular}%
}
\vspace{-8pt}
\end{wraptable}

We find that GPT-4o achieves the best performance as the base LLM. In contrast, reasoning models such as o3-mini and Deepseek-R1 perform slightly worse than GPT-4o, likely due to the relatively straightforward nature of visual reasoning in our tasks.

\section{Analysis}


\subsection{Comparison with Video-LLMs}

Compared with end-to-end Video-LLMs, the keyframe-based training-free method (e.g., \textsc{TASKER}) offers advantages in terms of training cost, inference efficiency, and interpretability. A detailed comparison with representative Video-LLMs, including performance and resource requirements, is provided in \textbf{Appendix~C}.

\subsection{Visualization}

\begin{figure*}[t]
  \centering
   \includegraphics[width=\linewidth]{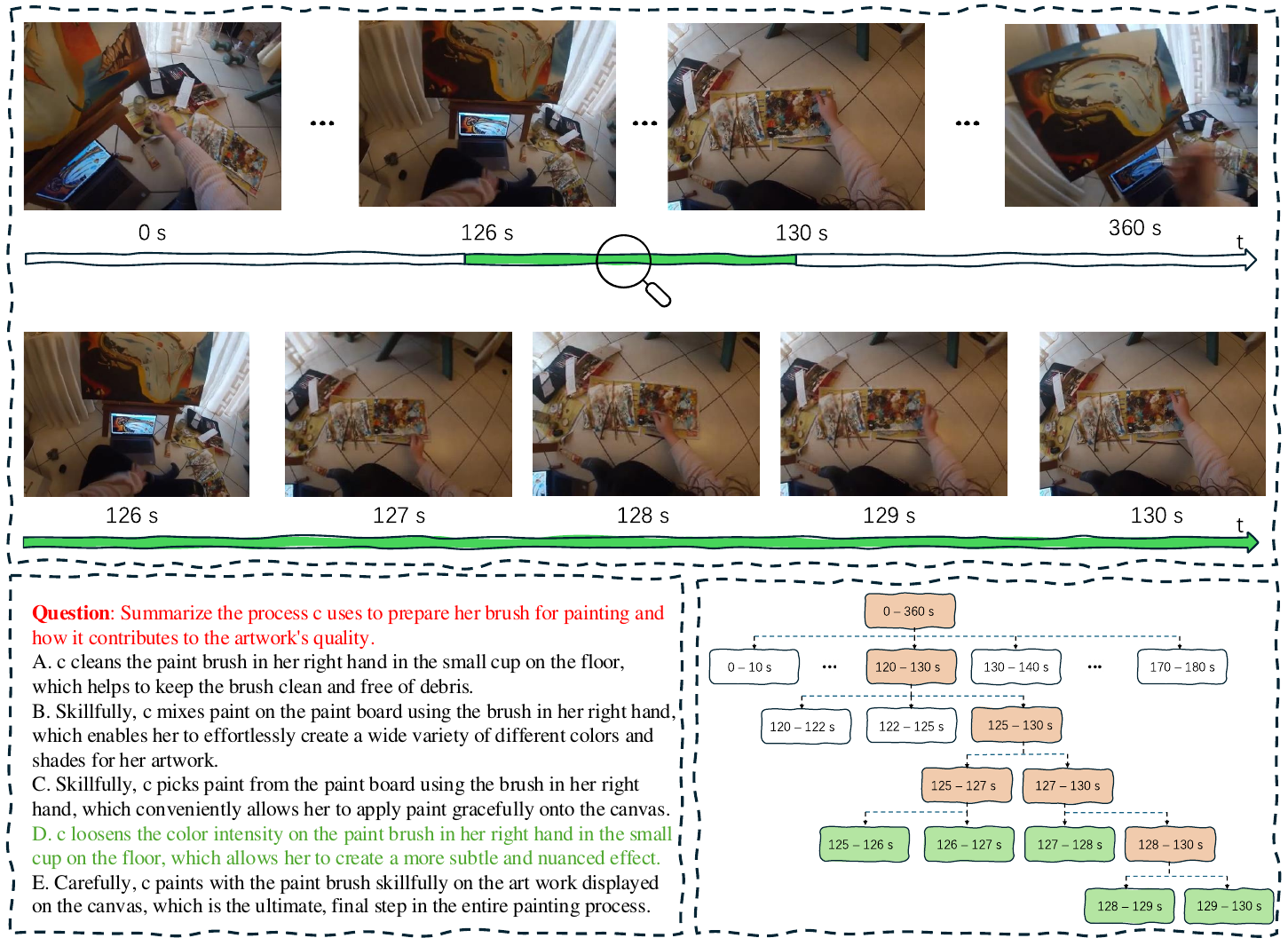}
   \caption{Visualization of tree-search and nodes expansion process of \textsc{TASKER} method solving a VideoQA case from EgoSchema~\cite{mangalam2023egoschemadiagnosticbenchmarklongform}.}
   \label{fig:visualization}
\end{figure*}


Figure \ref{fig:visualization} presents a visualized case study. In this 3-minute video, the key information for answering the question is located between 126s and 130s. Our \textsc{TASKER} algorithm precisely identifies this critical video segment by searching along the video tree and expanding relevant nodes. And it retrieves all frames within the 125s–130s range as keyframes, successfully answering the question. 
In the video tree, we mark the nodes traversed by the key search path in yellow and mark the final leaf node obtained from the key search path in green. The nodes outside the key search path are barely expanded.

\section{Conclusion}

In this work, we take a step toward deeper video understanding by connecting low-level VideoQA with higher-level video-guided agentic tasks. Specifically, we introduce \textsc{VG-GUI-Bench}, a benchmark that pairs tutorial videos with corresponding GUI interaction episodes to evaluate whether MLLM-based agents can extract procedural knowledge from videos and transfer it to long-horizon decision making. Building on the shared bottleneck of temporal content selection across both settings, we propose \textsc{TASKER}, a task-driven and scene-aware keyframe search algorithm that formulates keyframe extraction as a generalized graph-search problem, with MLLMs used to evaluate cost functions and termination confidence. Extensive experiments on VideoQA benchmarks demonstrate that \textsc{TASKER} consistently improves accuracy while using substantially fewer frames than prior keyframe-based baselines, highlighting a practical performance–efficiency trade-off in a training-free regime.

\section{Acknowledgement}

This work was supported by Fundamental and Interdisciplinary Disciplines Breakthrough Plan of the Ministry of Education of China (No. JYB2025XDXM101) and the National Natural Science Foundation of China (project No. 62495061, 62220106003).

\clearpage

\bibliographystyle{splncs04}
\bibliography{main}

\clearpage
\appendix
\renewcommand{\thesection}{\Alph{section}}
\setcounter{section}{0}

\begin{center}
    {\Large\bfseries Appendix}
\end{center}
\vspace{1em}

\section{Preliminary: Classic Search Algorithm}

Our \textsc{TASKER} algorithm is built on classic search algorithm in Algorithm \ref{algorithm_0}. Based on this fundamental process, the following search algorithms are distinguished by the method to determine priority for selecting nodes.

\begin{algorithm}[H]
\caption{Basic Search Algorithm}
\label{algorithm_0}
\begin{algorithmic}[1]
\Function{Search}{$\mathcal{N}_0$}
    \State Initialize open list $\mathcal{L} \gets \{\mathcal{N}_0\}$
    \While{$\mathcal{L}$ is not empty}
        \State $\mathcal{N} \gets$ Pop a node from $\mathcal{L}$ based on priority
        \If{$\mathcal{N}$ is the destination}
            \State \Return $\mathcal{N}$
        \EndIf
        \State Expand $\mathcal{N}$ to obtain neighboring nodes
        \State Add neighboring nodes to $\mathcal{L}$
    \EndWhile
    \State \Return None
\EndFunction
\end{algorithmic}
\end{algorithm}


\noindent
\textbf{Depth-First Search (DFS)} prioritizes nodes with greater depth and explores as far as possible
before backtracking.\\ 
\noindent
\textbf{Breadth-First Search (BFS)} explores all neighbors at the current level before moving on to the next level.\\
\noindent
\textbf{Greedy Best First Search (GBFS)} uses a heuristic evaluation function $h(n)$ as the cost function, i.e., $f(n) = h(n)$. Here, $h(n)$ represents the cost from the current node to the destination. It can guide the search algorithm towards the destination but does not guarantee an optimal path.\\
\noindent
\textbf{Dijkstra's Algorithm} uses a movement cost function $g(n)$ as the cost function, i.e., $f(n) = g(n)$. Here, $g(n)$ represents the cost of moving from the starting point to the current node. It finds the shortest path from a starting node to all other nodes by considering the weights of edges.\\
\noindent
\textbf{A* Algorithm} combines the benefits of Dijkstra's Algorithm and GBFS. The cost function is defined as: $f(n) = g(n) + h(n)$.
It balances efficiency and optimality, making it highly effective for path planning.

\section{Benchmarks}

{\bf EgoSchema}~\cite{mangalam2023egoschemadiagnosticbenchmarklongform}  dataset comprises over 5,000 human-curated multiple-choice question-answer pairs, making it one of the most widely used datasets for long-form video question and answering. Its subset contains 500 video and QA pairs. 
Each video in the datsset is three minutes in length. 
A notable feature of EgoSchema is its high difficulty level: humans can only achieve 76\% accuracy, and current Video-LLMs perform below 70\%. The extended video length and increased complexity underscore the importance of keyframe search and key information retrieval.

\noindent
{\bf NExT-QA}~\cite{xiao2021nextqanextphasequestionansweringexplaining} dataset consists of 5,440 videos and approximately 52K manually annotated question-answer pairs. Its primary focus is to assess whether QA models truly understand the causal and temporal structures of actions within a video.
We use the multiple-choice QA part of NExT-QA. Based on the types, the questions are divided into casual questions, temporal questions and descriptive questions. And based on the type of questions, it is divided into the following three categories.
\begin{itemize}
    \item \textbf{Causal Questions} seek to explain the causes or intentions behind actions, either by uncovering past motivations or predicting future outcomes; 
    \item \textbf{Temporal Questions} evaluate the model's ability to reason about the sequence and timing of actions, asking about past, present, or future events; 
    \item \textbf{Descriptive Questions} focus on detailing the scene by asking about places, objects, attributes, and key actions or events in the video.
\end{itemize}

\section{Comparison with Video-LLMs}

As previously discussed, there are two primary method for VideoQA task: (I) utilizing Video-LLMs for end-to-end computation; (II) employing (M)LLM-driven, keyframe-based, training-free method, like \textsc{TASKER}. We argue that both methods have their respective advantages. The key strength of Method I is that state-of-the-art Video-LLMs~\cite{gao2024linvtempowerimagelevellarge, chen2025expandingperformanceboundariesopensource} outperform Method II. It is suitable for scenarios where high accuracy is required, and computational cost is not a concern. In contrast, the primary advantage of Method II is its practical value for daily video analysis tasks, as it offers a more favorable balance between performance and computational cost. In the following, we use \textsc{TASKER} as an example to illustrate the relative advantages of Method II.

\begin{itemize}
    \item \textbf{Training-Free}. The training-free nature of Method II significantly reduces the overall cost. In Table \ref{table:compare_with_vlm}, we present the training costs and resource requirements of Video-LLMs that achieve comparable performance with our method on EgoSchema~\cite{mangalam2023egoschemadiagnosticbenchmarklongform} and NExT-QA~\cite{xiao2021nextqanextphasequestionansweringexplaining} benchmarks. The table highlights the complexity and high cost of training Video-LLMs, underscoring the training-free advantage of Method II.

    \item \textbf{Lower Inference Overhead}. 
    Method II still relies on large model inference. However, \textsc{TASKER} significantly reduces inference overhead by efficient keyframe selection instead of processing the whole video.

    \item \textbf{Better Interpretability}. 
    \textsc{TASKER} provides greater interpretability by generating intermediate results, such as the keyframe selection and textual reasoning process. In contrast to the end-to-end nature of Method I, this enhances transparency and interpretability. 
\end{itemize}

\begin{table*}[ht!]
\begin{center}
\caption{\textbf{Comparison of computation costs with Video-LLMs}}
\label{table:compare_with_vlm}
\resizebox{\linewidth}{!}{%
\begin{threeparttable}[b]
\setlength{\tabcolsep}{1.6mm}
\begin{tabular}{lccccc}
\toprule

\textbf{Model} & \textbf{Train Model Size} & \textbf{Tain Data Size} & \textbf{Computational Resources} & \textbf{EgoSchema} & \textbf{NExT-QA} \\

\midrule

ViLA~\cite{wang2024vilaefficientvideolanguagealignment} & 4B & 36.4K & 8 $\times$ 40 GB A100s & - & 75.6\ \\

VideoChat2~\cite{li2024mvbenchcomprehensivemultimodalvideo}& 7B & 4M & 32 $\times$ 80 GB A100s & 54.4 & 78.6  \\

VideoLLaMA2~\cite{cheng2024videollama2advancingspatialtemporal} & 72B & 13.6M & 32 $\times$ 80 GB A100s & 63.9 & 75.6 \\

\midrule

\textsc{TASKER} (ours) &  & Training-free &  & 63.1 & 77.4  \\

\bottomrule
\end{tabular}

\end{threeparttable}
}
\end{center}
\end{table*}

\section{Implementation Details}

For EgoSchema and NeXT-QA benchmarks, we first generate captions for the visible frames and then use LLMs to perform reasoning based on these captions. Following VideoAgent~\cite{wang2024videoagentlongformvideounderstanding}, we employ CogAgent~\cite{hong2024cogagentvisuallanguagemodel} as the captioner for the NExT-QA~\cite{xiao2021nextqanextphasequestionansweringexplaining} benchmark and LaViLa~\cite{zhao2022learningvideorepresentationslarge} for the EgoSchema~\cite{mangalam2023egoschemadiagnosticbenchmarklongform} benchmark due to its egocentric video pretraining.

The specific versions of the base LLMs we use are gpt-4-1106-preview and gpt-4o-2024-11-20. In \textsc{TASKER}, the number of visible frames can be adjusted by modifying the initial number of segments ($M$) and the maximum search iterations ($T$), which in turn affects the final accuracy. In our main experiments, we set $M = 10$ and $T = 6$.

\section{Detailed Demonstration of \textsc{VG-GUI-Bench}}

\begin{figure*}[htbp] 
  \centering
  \includegraphics[width=0.8\linewidth]{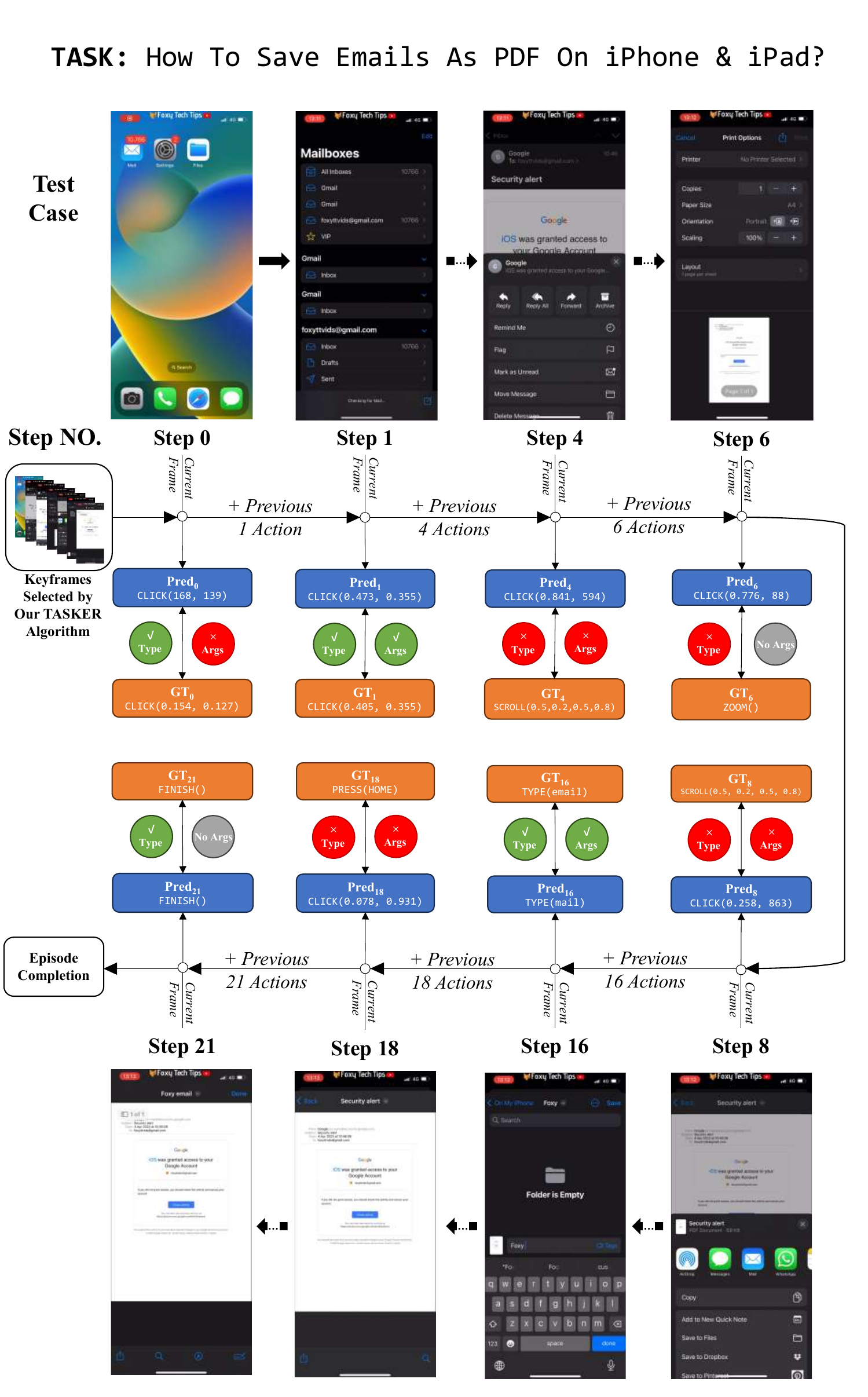} 
  \caption{A detailed demonstration of a test case from the \textsc{VG-GUI-Bench} benchmark. The example presents a multi-step task (i.e., saving emails as PDF on an iOS device), displaying the current GUI frame at each step alongside previous actions and the reference keyframes selected by our \textsc{TASKER} algorithm. It also visualizes the evaluation process by comparing the model's predicted actions (\textit{Pred}, including type and arguments) with the ground truth (\textit{GT}) until the completion of the episode.}
  \label{fig:appendix_benchmark_demo}
\end{figure*}

To provide a more intuitive understanding of our proposed \textsc{VG-GUI-Bench}, we present a detailed case study in Figure~\ref{fig:appendix_benchmark_demo}. This test case evaluates an agent's ability to execute a multi-step instruction (e.g., saving emails as PDF on an iOS device). 

As illustrated in the figure, the evaluation pipeline assesses the agent at each step of the interaction. At any given step, the input consists of the current GUI frame, previous actions, and the reference tutorial keyframes automatically extracted by our \textsc{TASKER} algorithm, which serve as essential visual guidance. Based on this input, the MLLM predicts the required action type (e.g., \texttt{CLICK}, \texttt{SCROLL}) and its corresponding arguments (e.g., screen coordinates). We then evaluate the performance by conducting a step-by-step comparison between the model's predicted actions (\textit{Pred}) and the human-annotated Ground Truth (\textit{GT}). This process iteratively continues until the completion of the episode (e.g., the model outputs a \texttt{FINISH()} action).

\section{Prompts}

\subsection{\textsc{TASKER} Keyframe Selector Prompts}

In the keyframe selection phase, we use the following prompts to guide the MLLM in evaluating candidate video segments and determining whether the current keyframes are sufficient. 
The prompt boxes below show the instructions for different search strategies (BFS, GBFS, Dijkstra, and A*), followed by the prompt for the self-evaluation (QA \& Reflect) step.

\begin{promptbox}{Prompt for BFS Strategy}
[System Prompt]
You are a highly strict UI navigation assistant designed to output JSON.

[User Prompt]
You are provided with sequential images sampled from a UI interaction video.
Each image is labeled with its frame index. The images are shown in chronological order.
Goal: {goal}

Candidate video segments (gaps between current frames):
{segment_des}

(BFS Strategy - Breadth Exploration)
Determine which segments are likely to contain crucial missing UI actions necessary to achieve the Goal. 
You are ALLOWED and ENCOURAGED to select MULTIPLE segments simultaneously if you believe important actions are missing in different video gaps.

Return JSON format exactly like this (can contain one or multiple items): 
{{"frame_descriptions": [
    {{"segment_id": "1", "duration": "xxx - xxx", "description": "Missing initial click"}}, 
    {{"segment_id": "3", "duration": "yyy - yyy", "description": "Missing confirmation popup"}}
]}}
\end{promptbox}

\begin{promptbox}{Prompt for GBFS Strategy (Focus on Missing Goal-Critical Actions)}
[System Prompt]
You are a highly strict UI navigation assistant designed to output JSON.

[User Prompt]
You are provided with sequential images sampled from a UI interaction video.
Each image is labeled with its frame index. The images are shown in chronological order.
Goal: {goal}

Candidate segments (gaps between current frames):
{segment_des}

(GBFS Strategy - Focus on Missing Goal-Critical Actions)
Determine which SINGLE segment is MOST LIKELY to contain the frames showing crucial missing UI actions needed to achieve the goal.
Focus on gaps in the operation flow: where does the current frame sequence fail to explain HOW the user got from one state to the next?

Return JSON format: {{"frame_descriptions": [{{"segment_id": "1", "duration": "xxx - xxx", "description": "Contains missing goal-critical actions"}}]}}
\end{promptbox}

\begin{promptbox}{Prompt for Dijkstra Strategy (Focus on UI State Changes)}
[System Prompt]
You are a highly strict UI navigation assistant designed to output JSON.

[User Prompt]
You are provided with sequential images sampled from a UI interaction video.
Each image is labeled with its frame index. The images are shown in chronological order.

Candidate segments (gaps between current frames):
{segment_des}

(Dijkstra Strategy - Focus on UI State Changes)
Identify which SINGLE candidate video segment contains the MOST significant UI state transition between its start frame and end frame.
Look at the actual images for the start and end frames of each segment to judge the visual difference.

In GUI operations, important UI state changes include (even if visually subtle):
- Page or screen navigation (e.g., moving from one view to another)
- Dialog boxes, pop-ups, or dropdown menus appearing or disappearing
- Button click effects or toggle state changes
- Text input or form field content changes
- Loading spinners, progress bars, or status indicator updates
- Sidebar, tab, or panel switching

Prioritize the segment where the start frame and end frame represent the MOST DISTINCT operational states, even if the pixel-level difference appears small.

Return JSON format: {{"frame_descriptions": [{{"segment_id": "1", "duration": "xxx - xxx", "description": "Most significant UI state transition"}}]}}
\end{promptbox}

\begin{promptbox}{Prompt for A* Strategy (Balance Info and State Changes)}
[System Prompt]
You are a highly strict UI navigation assistant designed to output JSON.

[User Prompt]
You are provided with sequential images sampled from a UI interaction video.
Each image is labeled with its frame index. The images are shown in chronological order.
Goal: {goal}

Candidate segments (gaps between current frames):
{segment_des}

(A* Strategy - Balance missing goal-relevant info and UI state changes)
Identify ONE single candidate segment that BEST satisfies BOTH conditions simultaneously:
1. GOAL PROXIMITY: The segment likely contains crucial missing UI actions that are necessary steps toward achieving the Goal. Without seeing the frames in this segment, the operation flow has an unexplained gap.
2. STATE CHANGE MAGNITUDE: Look at the start frame and end frame images of each segment. The segment whose boundary frames show the MOST different UI states is more likely to contain important operations.

In GUI operations, even subtle visual differences can represent critical steps (e.g., a single checkbox toggle, a dropdown selection, text typed into a field). Do NOT dismiss segments just because the visual change appears small - focus on whether an operational step is missing.

Return JSON format: {{"frame_descriptions": [{{"segment_id": "1", "duration": "xxx - xxx", "description": "Best A* candidate: missing goal step + UI state change"}}]}}
\end{promptbox}

\begin{promptbox}{Prompt for Self-Evaluation (QA \& Reflect)}
[System Prompt]
You are a highly strict UI navigation assistant designed to output JSON.

[User Prompt]
Task Goal: {goal}
Your sequential UI analysis: {answer_str}

As a strict UI automation tester, evaluate the VISUAL CONTINUITY of these frames. 
Can a user replicate this task step-by-step based on these frames?

Evaluate your confidence level strictly:
1: Severe Jumps (There are completely missing screens or sudden state changes. E.g., jumping from home to settings without seeing the menu. MUST expand.)
2: Minor Disconnects (The flow makes sense, but some button clicks, typing actions, or intermediate loading states are missing. Should expand.)
3: Strong Continuity (The frames capture all important UI actions and transitions. The operation flow is clear and a user can follow without confusion. Very minor intermediate states may be missing but no key step is skipped.)

CRITICAL RULE: UI tasks require high precision. If you have to GUESS what major action happened between any two frames, you MUST output 1 or 2. Do NOT output 3 unless the key action flow is clearly continuous with no important gaps.

Output JSON exactly like this: {{"confidence": 1}}
\end{promptbox}

\subsection{\textsc{VG-GUI-Bench} Evaluation Prompts}

For the evaluation on \textsc{VG-GUI-Bench}, the model is required to predict the next action. We define three different settings based on the provided visual context: No Video (baseline), Oracle Keyframe (ground truth guidance), and Selection Methods (keyframes extracted by algorithms like \textsc{TASKER}). The prompt boxes below detail the system and user prompts used in these settings.

\begin{promptbox}{Prompt for ``No Video'' Setting}
[System Prompt]
You are an expert GUI automation agent.

TASK OVERVIEW:
You are assisting a user to complete a task on a mobile device. You are given a single screenshot showing the current state of the device UI.

INPUT STRUCTURE:
You will receive exactly ONE image and text context (Goal & Previous Actions):
1. **Target Screen (The ONLY image):** This is the Current State of the device UI. This is the screen you must interact with.

YOUR REASONING PROCESS:
1. **Understand the goal:** Read the "Task Goal" to understand what the user is trying to accomplish.
2. **Locate your position:** Read the "Previous Actions" carefully. Determine which steps have already been completed and where you currently are in the overall workflow.
3. **Analyze the screen:** Examine the Target Screen thoroughly - identify all visible UI elements, buttons, text fields, menus, and interactive components.
4. **Infer the next step:** Based on the goal, previous actions, and the current screen state, reason about what the immediate next action should be.
5. **Extract coordinates:** Locate the exact UI element to interact with and derive its center coordinates.

### Output Format
You must output EXACTLY ONE action using the following function call syntax:

1. **CLICK(x, y)**
   - Tap at the specific normalized coordinate (0.0-1.0).
   - Example: `CLICK(0.53, 0.81)`

2. **SCROLL(x1, y1, x2, y2)**
   - Swipe from (x1, y1) to (x2, y2).
   - Example: `SCROLL(0.5, 0.8, 0.5, 0.2)`

3. **TYPE(content)**
   - Type text into the focused input field.
   - Example: `TYPE("Search query")`

4. **PRESS(key)**
   - Keys: "BACK", "HOME", "ENTER".
   - Example: `PRESS("BACK")`

5. **ZOOM()**
   - Perform a zoom or multi-touch gesture.

6. **FINISH()**
   - The task is completed or impossible to continue.

### Strict Rules
- The coordinates MUST pinpoint the center of the target UI element on the Target Screen.
- Coordinates must be normalized (0.0 to 1.0).
- Output NOTHING ELSE but the function call string. No explanations."""

[User Prompt]
Task Goal: {}

Previous Actions: {}

Based on the Task Goal, Previous Actions, and the current Target Screen, what is the exact Next Action? Output ONLY the action format:
\end{promptbox}

\begin{promptbox}{Prompt for ``Oracle Keyframe'' Setting}
[System Prompt]
You are a GUI automation agent.

YOUR TASK IS VISUAL EXTRACTION:
You have been provided with a sequence of images.
1. The **VERY FIRST** image is the "Ground Truth" key (Target State), where the correct element is explicitly marked with a RED BOX.
2. The **LAST** image is the raw input (Current State) where you need to perform the action.
3. The images **BETWEEN** the first and the last are random examples/noise. **IGNORE THEM COMPLETELY.**

CRITICAL INSTRUCTION:
1. FOCUS ONLY on the **FIRST** image. Locate the precise center (x, y) of the RED BOX.
2. IGNORE any red boxes in the intermediate images.
3. Map the coordinates from the FIRST image directly to the LAST image as your `CLICK(x, y)`.
4. Do not perform semantic reasoning. Trust the RED BOX in the FIRST image implicitly.

### Output Format
You must output EXACTLY ONE action using the following function call syntax:

1. **CLICK(x, y)**
  - Tap at the specific normalized coordinate (0.0-1.0).
  - Example: `CLICK(0.53, 0.81)`

2. **SCROLL(x1, y1, x2, y2)**
  - Perform a drag/swipe gesture.
  - Example: `SCROLL(0.5, 0.8, 0.5, 0.2)`

3. **TYPE(content)**
  - Type text.
  - Example: `TYPE("Hello World")`

4. **PRESS(key)**
  - Keys: "BACK", "HOME", "ENTER".
  - Example: `PRESS("BACK")`

5. **ZOOM()**
  - Perform a zoom or multi-touch gesture.

6. **FINISH()**
  - The task is completed or impossible to continue.

### Rules
- Output ONLY the function call string.
- Coordinates must be normalized (0.0 to 1.0).

[User Prompt]
Instruction: {}

Previous Actions: {}

Output the Next Action (Extract from the first image):
\end{promptbox}

\begin{promptbox}{Prompt for ``Selection Methods'' Setting (e.g. TASKER)}
[System Prompt]
You are an elite UI automation agent.

INPUT STRUCTURE:
You will receive a sequence of images:
1. The images before the last one are "Golden Keyframes" dynamically extracted from a video tutorial. They demonstrate the EXACT, flawless visual sequence to successfully complete the user's overarching goal.
2. The VERY LAST IMAGE is the "Target Screen" (Current State). This is the ONLY screen you will interact with.

YOUR MISSION TO SUCCEED:
1. Analyze the "Golden Keyframes" to deeply understand the workflow, the visual changes, and the ultimate intent.
2. Read the "Previous Actions" provided in the prompt to know exactly where you are in the sequence.
3. Determine the EXACT NEXT ACTION required to progress the task, using the Golden Keyframes as your visual cheat sheet.
4. Derive the coordinates (x, y) STRICTLY from the VERY LAST IMAGE (Target Screen).

### Output Format
You must output EXACTLY ONE action using the following function call syntax:

1. **CLICK(x, y)**
  - Tap at the specific normalized coordinate (0.0-1.0).
  - Example: `CLICK(0.53, 0.81)`

2. **SCROLL(x1, y1, x2, y2)**
  - Example: `SCROLL(0.5, 0.8, 0.5, 0.2)`

3. **TYPE(content)**
  - Example: `TYPE("Search query")`

4. **PRESS(key)**
  - Keys: "BACK", "HOME", "ENTER".
  - Example: `PRESS("BACK")`

5. **ZOOM()**
  - Perform a zoom or multi-touch gesture.

6. **FINISH()**
  - The task is completed or impossible to continue.

### Strict Rules
- The coordinates MUST pinpoint the center of the actionable UI element on the Target Screen.
- Output NOTHING ELSE but the function call string. No explanations."""

[User Prompt]
Task Goal: {}

Previous Actions: {}

Based on the provided Reference Frames and the Previous Actions, what is the exact Next Action to perform on the VERY LAST IMAGE (Target Screen)? Output ONLY the action format:
\end{promptbox}

\section{Results on OSWorld}

OSWorld~\cite{xie2024osworld} are not video-guided by design, so it does not directly match \textsc{TASKER}'s target setting. Still, we collected instructional videos for a subset of OSWorld tasks and evaluated Gemini-3-Flash with no video, uniform-sampled video frames, and TASKER-selected video frames. Table~\ref{tab:osworld_advantage} shows that TASKER improves overall performance, though gains vary by domain.

\begin{table}[h]
\centering
\small
\caption{\textbf{Task Success Rate of Gemini-3-Flash on OSWorld Subset}. The best performances are highlighted in bold.}
\label{tab:osworld_advantage}
\resizebox{\linewidth}{!}{
\begin{tabular}{l|cccccc}
\toprule
Method & Overall & Impress & Writer & OS & ThBrd & VSCode \\
\midrule
No Video
& 46.27 & 32.76 & \textbf{46.65} & \textbf{46.67} & \textbf{64.29} & 38.46 \\
Uniform Sample Frames
& 49.13 & 49.19 & \textbf{46.65} & \textbf{46.67} & 57.14 & 46.15 \\
TASKER (ours)
& \textbf{50.63} & \textbf{49.49} & 39.98 & 40.00 & \textbf{64.29} & \textbf{61.54} \\
\bottomrule
\end{tabular}
}
\end{table}

\end{document}